\documentclass[11pt]{article}

\usepackage[preprint]{acl}

\usepackage{times}
\usepackage{latexsym}
\usepackage{float}
\usepackage[T1]{fontenc}
\usepackage[utf8]{inputenc}

\usepackage{microtype}

\usepackage{inconsolata}

\usepackage{graphicx}

\usepackage{amsmath}
\usepackage{booktabs}
\usepackage{multirow}
\usepackage{colortbl}
\usepackage{xcolor}
\usepackage{newunicodechar}
\usepackage{enumitem}

\newunicodechar{（}{(}
\newunicodechar{）}{)}
\definecolor{gray94}{gray}{0.94}

\title{Why Users Go There: World Knowledge-Augmented Generative \\ Next POI Recommendation}

\author{
 \textbf{Qiuyu Ding\textsuperscript{1}},
 \textbf{Heng-Da Xu\textsuperscript{1}},
 \textbf{Wei Zhang\textsuperscript{1}},
 \textbf{Dongyi Lv\textsuperscript{1,2}},
 \textbf{Changda	Xia\textsuperscript{1}},
\\
 \textbf{Feng Xiong\textsuperscript{1}},
 \textbf{Mu Xu\textsuperscript{1}\thanks{Corresponding author.}}, 
 \vspace{2pt}
\\
 \textsuperscript{1}Amap, Alibaba Group \ 
 \textsuperscript{2}Xi’an Jiaotong University,
 \vspace{2pt}
\\
    \texttt{\{qiuyu.dqy,xuhengda.xhd,chengxin.zw,xiachangda.xcd,xf250971,xumu.xm\}@alibaba-inc.com} \\
    \texttt{lvdongyi@stu.xjtu.edu.cn}
}

\begin{document}
\maketitle
\begin{abstract}

Generative point-of-interest (POI) recommendation models based on large language models (LLMs) have shown promising results by formulating next POI prediction as a sequence generation task. However, the knowledge encoded in these models remains fixed after training, making them unable to perceive evolving real-world conditions that shape user mobility decisions, such as local events and cultural trends. To bridge this gap, we propose AWARE (Agent-based World knowledge Augmented REcommendation), which employs an LLM agent to generate location- and time-aware contextual narratives that capture regional cultural
characteristics, seasonal trends, and ongoing events relevant to each user. Rather than introducing generic or noisy information, AWARE further anchors these narratives in each user's behavioral context, grounding external world knowledge in personalized spatial-temporal patterns. Extensive experiments on three real-world datasets demonstrate that AWARE consistently outperforms competitive baselines, achieving up to 12.4\% relative improvement. 
% Further analysis confirms that world knowledge is effective only when sufficiently grounded in user-specific behavior, yielding complementary signals that capture context-dependent intent unobservable from check-in data alone.
\end{abstract}

\section{Introduction}
\label{sec:intro}
With the proliferation of location-based social networks (LBSNs) and mobile devices, massive volumes of user check-in data have become available, enabling the study of human mobility at an unprecedented scale~\cite{gonzalez2008understanding}. A central task in this domain is \textit{next point-of-interest (POI) recommendation}: given a user's historical check-in trajectory, predicting the most likely next location they will visit~\cite{li2024large,feng2024move, halder2022efficient, zhong2025comapoi}. Accurate next POI prediction directly benefits real-world applications, including personalized navigation, location-aware advertising, and urban planning~\cite{zheng2015trajectory}, and has attracted increasing attention from both academia and industry in recent years. 

\begin{figure}[t]
  \centering
  \includegraphics[width=\linewidth]{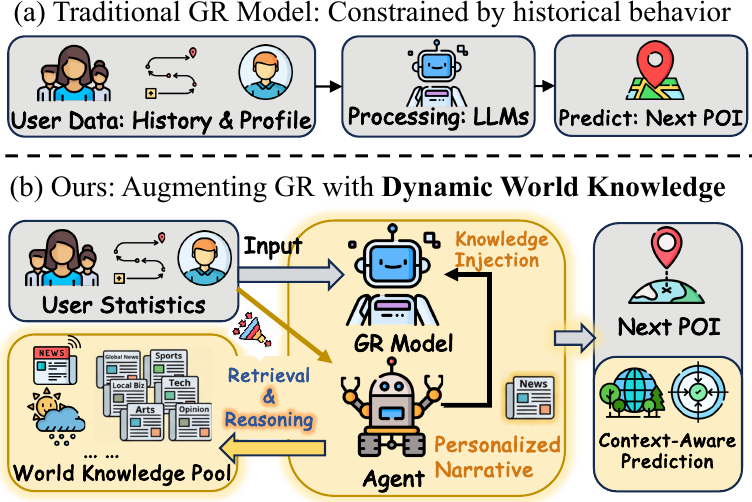}
  \caption{Comparison of (a) traditional generative recommendation using only historical check-in data, and (b) AWARE, which augments the GR model with dynamic world knowledge selected by an LLM agent.}
  \label{fig:intro}
\end{figure}

Existing approaches have evolved from collaborative filtering and Markov chain models~\cite{feng2015personalized} to recurrent and attention-based architectures~\cite{hochreiter1997long,luo2021stan,yang2022getnext,yan2023spatio} that capture increasingly complex spatial-temporal dependencies. More recently, large language models (LLMs) \cite{ding2026inductive,rajput2023recommender,chen2025onesearch,jiang2026llm,zhou2025onerec} have given rise to \textit{generative POI recommendation}~\cite{li2024large,wang2025generative,liu2025geography,ROS}, which serializes check-in history into a natural language prompt and fine-tunes an LLM to directly generate the next POI. By exploiting rich textual representations rather than opaque ID embeddings, generative approaches achieve strong performance on next POI prediction benchmarks. 

Despite their promise, existing generative POI recommendation models share a critical limitation: once fine-tuned on historical data, they operate as \textbf{static systems} whose knowledge is entirely frozen at training time. This renders them unable to capture evolving real-world context, such as local events, cultural trends, and seasonal shifts, that fundamentally shapes user mobility decisions. For example, a user who routinely visits a gym on weekday evenings may instead attend a community fitness event at a nearby park; a static model, unaware of this event, would confidently predict the gym and fail. Without access to such external context, existing methods capture \textit{where} users have been but not \textit{why} they go next. 

% To address this gap, we propose \textbf{AWARE} (\textbf{A}gent-based \textbf{W}orld knowledge \textbf{A}ugmented \textbf{RE}commendation), a framework that injects dynamic, personalized world knowledge into the generative recommendation pipeline (illustrated in Figure~\ref{fig:intro}). Specifically, for each user's activity region and time period, AWARE employs an LLM agent to synthesize contextual narratives, termed \textit{hotspot text}, describing regional cultural characteristics, seasonal trends, and ongoing events. Unlike retrieval-augmented generation (RAG), which returns pre-indexed documents irrespective of user identity, our agent reasons over each user's mobility history to produce personalized narratives that distill relevant world context into actionable conclusions. However, this generative flexibility also introduces a key challenge: grounding the world knowledge in individual user context rather than introducing generic or noisy information. Accordingly, AWARE anchors the hotspot text in signals derived from each user's mobility history, such as frequently visited locations and habitual spatial transitions, so that the injected world knowledge reflects individual behavioral regularities instead of undifferentiated context.

To address this gap, we propose \textbf{AWARE} (\textbf{A}gent-based \textbf{W}orld-\textbf{A}ugmented \textbf{RE}commendation), a framework that injects dynamic, personalized world knowledge into the
  generative recommendation pipeline (Figure~\ref{fig:intro}). For each user's activity region and time period, AWARE employs an LLM agent to synthesize contextual narratives, termed \textit{hotspot text},
  describing regional cultural characteristics, seasonal trends, and ongoing events. Unlike RAG, which returns pre-indexed documents irrespective of user identity, our agent reasons over each user's mobility history to produce personalized narratives that distill relevant world context into actionable conclusions. To ensure the injected knowledge remains user-specific rather than generic, AWARE further anchors the
  hotspot text in behavioral signals such as frequently visited locations and habitual spatial transitions.
% We summarize our contributions as follows:
% \begin{itemize}
%     \item We propose AWARE, a framework for next POI recommendation that injects dynamic world knowledge into the generative recommendation pipeline via an LLM agent, enabling the model to perceive real-world context such as local events and cultural trends beyond check-in data.
%     \item A user-knowledge alignment module is designed to ground world knowledge in individual user interests through POI visit frequency statistics and spatial transition patterns, ensuring only personally relevant knowledge is incorporated.
%     \item Extensive experiments on three real-world datasets demonstrate that AWARE consistently outperforms competitive baselines. Ablation studies and case studies further validate the effectiveness of each proposed component.
% \end{itemize}

% The contributions of this work can be summarized in three aspects: (1) we propose AWARE, which injects dynamic world knowledge into generative POI recommendation via an LLM agent, enabling the model to perceive real-world context beyond check-in data; (2) a grounding mechanism anchors world knowledge in each user's behavioral context through visit frequency and spatial transition patterns; and (3) extensive experiments on three datasets demonstrate consistent improvements over competitive baselines, with ablation and case studies validating each component.

%%%%%%%%%% Section Split %%%%%%%%%%

\section{Background}

\subsection{Problem Formulation}

Let $\mathcal{U}$ denote the set of users and $\mathcal{P}$ the set of POIs. Each POI $p \in \mathcal{P}$ is associated with a name, a category, and geographic coordinates. A check-in $(u, p, t)$ records that user $u$ visited POI $p$ at time $t$. For each user $u$, the check-in history is an ordered sequence $\mathcal{S}_u = [(p_1, t_1), (p_2, t_2), \ldots, (p_n, t_n)]$. The task is to predict the next POI $p_{n+1}$ given $\mathcal{S}_u$ and the current time context.

\begin{figure*}[t]
  \centering
  \includegraphics[width=\linewidth]{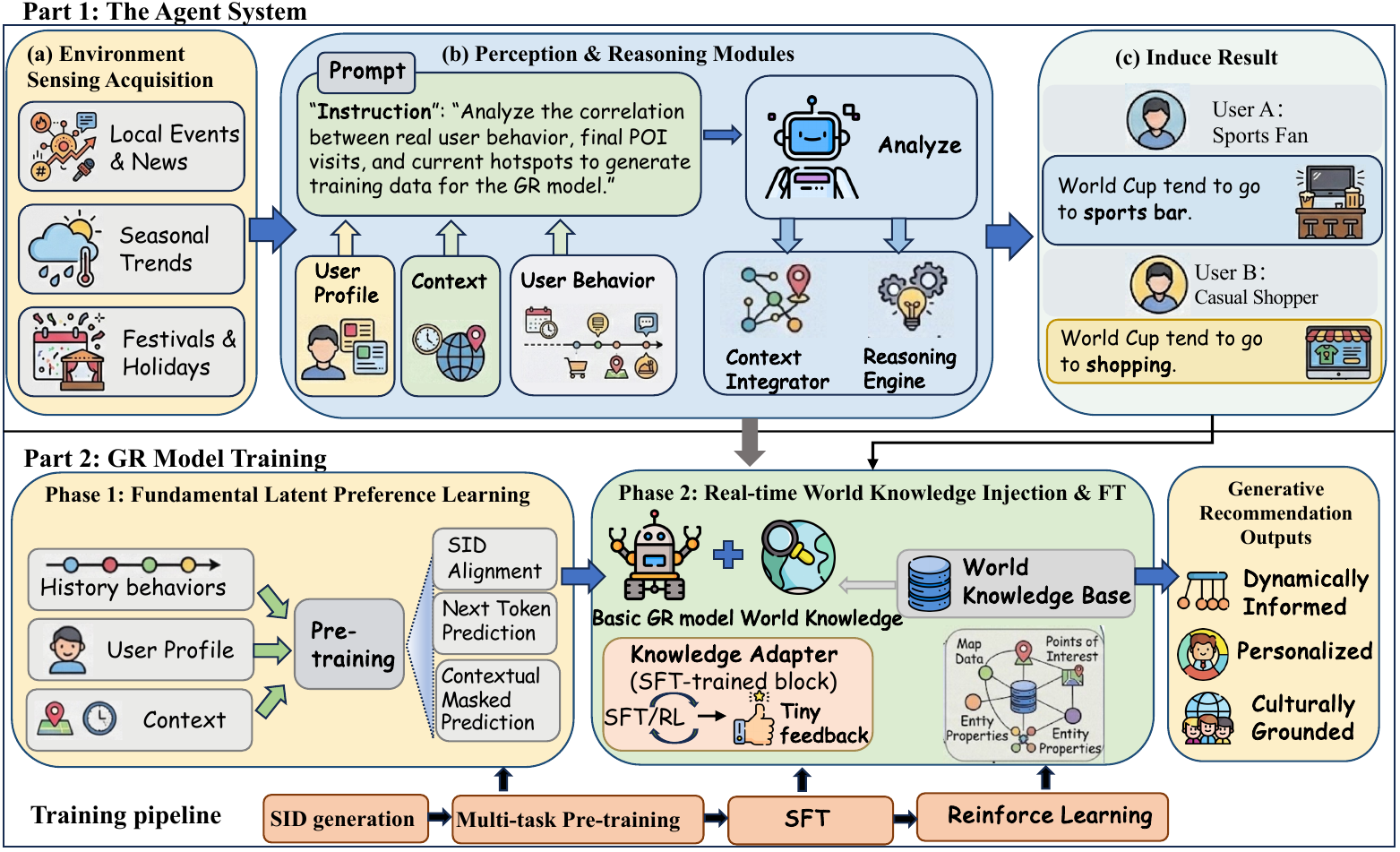}
  \caption{Overview of the AWARE framework. (1)~An LLM agent retrieves external signals (e.g., news, events, festivals) and integrates them with user behavioral patterns to produce personalized narratives. (2)~The pretrained GR model is fine-tuned with the acquired world knowledge to generate context-aware recommendations.}
  % \Description{Overview of the AWARE framework.}
  \label{fig:framework}
\end{figure*}

\subsection{Generative POI Recommendation}

Recent work reframes next POI prediction as a text generation task~\cite{li2024large,wang2025generative,liu2025geography,ROS}. A user's check-in history is serialized into a structured prompt containing POI names, categories, timestamps, and geographic information. An LLM is then fine-tuned to generate the next POI as its output sequence.

Formally, a prompt construction function $\phi(\cdot)$ converts $\mathcal{S}_u$ into a textual input $x = \phi(\mathcal{S}_u)$, and the model is trained to maximize $P(y \mid x)$, where $y$ is the ground-truth next POI. This formulation enables the model to leverage pre-trained semantic knowledge about location names and categories, providing richer representations than ID-based.

\subsection{Semantic ID}

Rather than representing POIs by raw names, which consume lengthy token sequences and lack structural relationships, recent work assigns each POI a compact \textit{Semantic ID} (SID)~\cite{wang2025generative}. SID encodes each POI as a short compositional token sequence whose components capture different facets and can be shared across POIs:

\begin{table}[ht]
    \centering
    \small % 控制字体大小
    \setlength{\tabcolsep}{4pt} % 调整列间距，让布局更紧凑
    
    \begin{tabular}{lcl} % 注意这里去掉了竖线符号 |
    % 注意：这里去掉了所有的 \hline 或 \toprule 等画线命令
    
    % 第一行
    \texttt{Coffee Shop, 1128 3rd Ave} & $\Rightarrow$ & \texttt{<a\_29><b\_31><c\_20>} \\
    % 第二行
    \texttt{Coffee Shop, 5 W 21st St} & $\Rightarrow$ & \texttt{<a\_26><b\_3><c\_20>} \\
    
    \end{tabular}
    \end{table}    

\noindent Both POIs share \texttt{<c\_20>} due to their common category, while the remaining tokens capture individual identity. AWARE is agnostic to the POI representation scheme; in this work we adopt a
SID-equipped generative backbone (details in Section~\ref{sec:method}).

\subsection{World Knowledge for POI Recommendation}

In this work, world knowledge refers to dynamic, location-specific contextual information beyond check-in records, including regional cultural characteristics, local events, and seasonal trends. Such knowledge evolves continuously and often drives user mobility decisions, yet cannot be encoded statically within model parameters.

We define world knowledge as $ w = \mathcal{A}(\mathcal{S}_u, l, t) $, where $\mathcal{A}$ denotes an LLM agent, $l$ is the user's activity region, and $t$ is a temporal context. Unlike generic knowledge retrieval, $\mathcal{A}$ first analyzes the user's mobility patterns from $\mathcal{S}_u$, then incorporates real-world events and trends relevant to region $l$ at time $t$. For example, given a user who commutes between Queens and Midtown Manhattan and has visited Yankee Stadium on weekends during April--May 2012, the agent may generate:

\vspace{0.5em}
\noindent\fbox{\parbox{0.95\linewidth}{\small\textit{``The user exhibits a commuting pattern between Queens and Midtown Manhattan. Combined with the Yankees' home season in April--May, the user is likely interested in upcoming home games or nearby venues.''}}}
\vspace{0.5em}

\noindent We term this output \textit{hotspot text}, a personalized narrative that bridges user behavior with external world dynamics to reason about why a user is likely to visit certain POIs.

\section{Methodology}
\label{sec:method}

In this section, we present the AWARE framework (see Figure~\ref{fig:framework}). We first describe the personalized world knowledge acquisition (Section~\ref{sec:wk}), then introduce user-aware knowledge alignment (Section~\ref{sec:grounding}), and finally detail the context-aware generative inference (Section~\ref{sec:generation}).

\subsection{World Knowledge Acquisition Agent}
\label{sec:wk}

% As discussed in Section~\ref{sec:intro}, existing generative POI models treat knowledge as static, with all contextual understanding frozen at training time. Yet user mobility is shaped by evolving real-world dynamics that no fixed-parameter model can capture. A natural remedy is to incorporate external sources such as news feeds, event calendars, and travel portals; however, such information is vast and largely irrelevant to any individual user, making direct injection impractical. This observation underpins AWARE's design principle: leveraging each user's behavioral patterns as an anchor for world knowledge selection, so that the distilled signals capture not only \textit{where} users have been but \textit{why} they are likely to go next.

Existing generative POI models treat knowledge as static, yet user mobility is shaped by evolving real-world dynamics that no fixed-parameter model can capture. External sources such as news feeds, event calendars, and travel portals offer a natural remedy, but their content is vast and largely irrelevant to any individual user. This motivates AWARE's design principle: leveraging each user's behavioral patterns as an anchor for world knowledge selection, so that the distilled signals capture not only \textit{where} users have been but \textit{why} they are likely to go next.

\subsubsection{Three-Stage Reasoning Strategy}
To this end, AWARE instantiates a tool-augmented LLM agent $\mathcal{A}$ that, given a user's trajectory $\mathcal{S}_u$ and the temporal--spatial context $(t, l)$ implied by the tail of the trajectory, is equipped with a retrieval toolbox $\mathcal{R} = \{\texttt{web\_search},\, \texttt{web\_fetch}\}$ and produces a compact hotspot text:
\begin{equation}
    w = \mathcal{A}\bigl(\mathcal{S}_u,\, t,\, l\,;\, \mathcal{R},\, \Pi\bigr),
\end{equation}
where $\Pi$ denotes the system prompt that governs the agent's reasoning protocol. The protocol decomposes into three sequential stages, with a detailed description provided below. 

\paragraph{Stage~(i): Behavioral Profile Inference.} The agent first parses $\mathcal{S}_u$ in chronological order and weights the tail of the sequence more heavily to identify the user's active neighborhood $l$, recurring interests (e.g., dining style, commute routine, fitness habits, nightlife preferences), and plausible near-term intent. This profile is retained as an internal anchor and never exposed verbatim in the final output, ensuring that downstream retrieval is driven by user-specific priors rather than generic demographic assumptions.

\paragraph{Stage~(ii): Constrained External Retrieval.} Conditioned on the inferred profile, the agent formulates targeted web queries that jointly bind three constraints: \textit{geographic} (restricted to $l$ and its surrounding neighborhoods), \textit{temporal} (anchored at $t$, with the target year and month embedded explicitly in every query), and \textit{thematic} (the inferred interest categories). Retrieval proceeds iteratively: broader \texttt{web\_search} calls first surface candidate event calendars, neighborhood guides, and news items, after which \texttt{web\_fetch} verifies specific claims on the most informative pages. The agent is explicitly instructed to corroborate each substantive claim across at least two independent sources and to discard any signal that cannot be anchored to the target time window, which mitigates hallucinations inherited from the LLM's parametric memory.

\paragraph{Stage~(iii): Grounded Synthesis.} In the final stage, the agent weaves the internal behavioral anchor together with the verified external signals into a single coherent paragraph $w$. The paragraph is required to interleave three elements in a tightly integrated manner: (a)~what the user's history suggests, (b)~what regional context is salient at time $t$, and (c)~what POI categories or venue styles the user is most likely to seek next. Outputs that merely recite the trajectory, or that read as generic city guides disconnected from user behavior, are rejected by the format constraints in $\Pi$, ensuring that $w$ reads as a \textit{user-grounded contextual interpretation} rather than a document summary.

\subsubsection{Agent Instantiation and Prompt Design}
\label{sec:agent-inst}
We instantiate $\mathcal{A}$ with a large-capacity LLM (\texttt{Qwen3-max}) augmented with a web-retrieval toolbox $\mathcal{R}$ that provides date-filtered web search and page-level content extraction. The system prompt $\Pi$ encodes the three-stage protocol as explicit instructions and imposes three key constraints: (i)~all factual claims must be grounded in retrieved evidence rather than parametric memory, (ii)~the output must be a single coherent paragraph, and (iii)~the length must not exceed a word budget $M$, which can be adjusted according to application requirements. Figure~\ref{fig:agent-prompt} visualizes the key components of $\Pi$, including the quality safeguards elaborated in Section~\ref{sec:runtime-guarantees}. The complete system prompt is provided in Appendix~\ref{app:agent-prompt}.

\begin{figure}[t]
    \small
    \setlength{\tabcolsep}{7pt}
    \renewcommand{\arraystretch}{1.15}
    \centering
    \begin{tabular}{|p{0.92\linewidth}|}
    \hline
    \rowcolor{blue!70!black}\multicolumn{1}{|l|}{\color{white}\textbf{System Prompt $\Pi$ of the Knowledge Agent}} \\
    \hline
    \textbf{Role.} The agent for next POI recommendation in city $l$. \\[0.3em]
    \textbf{Input.} User trajectory $\mathcal{S}_u$: \{(time, POI, type, address, lat, lon)\}. \\[0.3em]
    \textbf{Tools.} \texttt{web\_search} (with publication-date filter) and \texttt{web\_fetch}. \\[0.3em]
    \textbf{Protocol.} \\
    \quad (1) Infer user preferences, routines, and near-term intent from $\mathcal{S}_u$ (tail-weighted). \\
    \quad (2) Identify the active region and activity context. \\
    \quad (3) Issue parallel queries for region- and time-local hotspots, events, and trends. \\
    \quad (4) Synthesize one paragraph weaving behavior, context, and predicted POI categories. \\[0.3em]
    \textbf{Temporal fidelity.} Treat the latest timestamp in $\mathcal{S}_u$ as ``now''; embed the target year/month in every query; prefer archival sources; drop unverifiable claims. \\[0.3em]
    \textbf{Search budget.} At most two retrieval rounds; emit parallel queries rather than serial refinements. \\[0.3em]
    \textbf{Output.} One compact English paragraph ($\le M$ words); no structured formatting or trajectory recitation. \\
    \hline
    \end{tabular}
    \caption{Condensed view of the system prompt $\Pi$. Beyond the reasoning protocol, $\Pi$ encodes runtime guarantees for temporal fidelity, search-budget control, and output-length control.}
    \label{fig:agent-prompt}
\end{figure}

\subsubsection{Generation Quality Control}
\label{sec:runtime-guarantees}

Beyond the reasoning protocol, AWARE enforces three operational constraints on agent outputs, namely \textit{output-format control}, \textit{tool-usage efficiency}, and \textit{temporal fidelity}, each implemented across prompt, tool, and post-hoc validation layers to forestall any single failure mode.

\paragraph{Output-Format Control.} Since the downstream GR model is prompt-length sensitive, we enforce the word budget $M$ via a prompt, validator, and truncator cascade: $\Pi$ instructs the agent to cap $w$ at $M$ words and self-check length; if the validator detects $|w| > M$, the agent is re-invoked with a brief rewrite instruction; any residual overshoot is hard-truncated. We instantiate two budgets, $M=80$ and $M=150$, which anchor the informativeness versus noise trade-off in Section~\ref{sec:hotspot-length}.

\paragraph{Tool Usage Efficiency.} Since agent rollouts dominate preprocessing cost, $\Pi$ caps retrieval at two rounds, a broad main round and an optional supplemental round for residual gaps, and encourages parallel rather than sequential queries within each round; our orchestrator dispatches them concurrently. A code-level guard hard-stops any attempt at a third round and forces synthesis, bounding per-user latency and cost.

\paragraph{Temporal Fidelity.} As the evaluation benchmarks span 2009 to 2013 and predate contemporary LLM training corpora, AWARE grounds the agent in the user's own timeline: the latest timestamp in $\mathcal{S}_u$ is treated as ``now,'' every query is anchored to the target year and month, and documents lacking period-appropriate corroboration are discarded, ensuring that the injected knowledge reflects the user's original temporal and spatial context.

\subsection{User-Aware Knowledge Alignment}
\label{sec:grounding}

Indiscriminate injection of world knowledge risks introducing noise. To ensure that external context is actionable for each user, AWARE aligns world knowledge with user-specific behavioral evidence through three complementary priors.

\noindent \textbf{Frequency-Based Behavioral Prior.} We compute a frequency-ranked list of each user's visited POIs: 
\begin{equation}
    \mathcal{F}_u = [(p_1, c_1, f_1), (p_2, c_2, f_2), \ldots]
\end{equation}
where $p_i$ denotes a POI, $c_i$ its category, and $f_i$ the visit count. This prior encodes long-term behavioral regularities and serves as a relevance filter: signals aligning with established patterns
receive implicit reinforcement, while those lacking behavioral corroboration are down-weighted.

\noindent \textbf{Transition-Based Spatial Grounding.} We model user mobility dynamics by extracting POI-to-POI transition statistics. For each POI $p_i$ visited by user $u$, we record the set of subsequent destinations and their frequencies:
\begin{equation}
    \mathcal{T}_u(p_i) = \{(p_j, n_{ij})\}_{j=1}^{|\mathcal{N}(p_i)|}
\end{equation}
where $p_j$ is a destination POI and $n_{ij}$ the transition count. Given the most recently visited POI, we select the top-$k$ most frequent transitions as spatial grounding signals, channeling world knowledge
  through the user's movement structure. 

\noindent \textbf{Periodic Temporal Modeling.} User mobility often exhibits strong \textit{periodicity}: behavior on Monday mornings is more predictive of next Monday than activity from the preceding weekend. To capture such long-range periodic evidence, AWARE reserves a fraction $\beta$ of the history budget for entries sharing the same day-of-week as the prediction target, filling the remaining slots with the most recent entries. Formally, given budget $L$ and target day-of-week $d$, we select $\lfloor \beta L \rfloor$ entries from $\{e \in \mathcal{S}_u \mid \text{dow}(e) = d\}$ and $(L - \lfloor \beta L \rfloor)$ recent entries from the remainder. This ensures that temporal knowledge from the agent (e.g., ``Friday evening trends'') is paired with the user's corresponding historical behavior.

\subsection{Context-Aware Generative Inference}
\label{sec:generation}

% Before describing how AWARE constructs its input, we briefly outline the generative recommendation pipeline that serves as the foundation. Each POI is assigned an SID by embedding its textual description through a pretrained language model and discretizing the result via hierarchical clustering into a compositional token sequence. Building upon the base SID, a geospatial prefix derived from S2 cells~\cite{s2geometry} is prepended to encode geographic locality at multiple resolutions, enabling POIs in spatial proximity to share prefix tokens. During training, each sample consists of a user's historical check-in sequence, where each entry is represented by its timestamp and the corresponding POI's SID, paired with the ground-truth next POI's SID as the generation target \cite{ROS}. At inference time, the model autoregressively generates the SID token sequence of the predicted next POI. Notably, AWARE introduces world knowledge and behavioral priors entirely at the representation level, requiring no architectural modification to the generative backbone, which makes it readily applicable to other SID-based or generative recommendation frameworks.

Before describing how AWARE constructs its input, we briefly outline the generative recommendation (GR) pipeline. Each POI is assigned an SID by embedding its textual description through a pretrained language model and discretizing via hierarchical clustering. A geospatial prefix derived from S2 cells~\cite{s2geometry} is prepended to encode geographic locality at multiple resolutions, enabling spatially proximate POIs to share prefix tokens. The model is trained to generate the ground-truth next POI's SID given the user's historical sequence, and autoregressively produces the predicted SID at inference time~\cite{ROS}.

Building on this pipeline, AWARE organizes the acquired knowledge (Sec.~\ref{sec:grounding}) into a structured prompt providing three complementary dimensions beyond the check-in sequence:
\begin{equation}
    x = \phi(\mathcal{H}_u, \mathcal{F}_u, w, \mathcal{C}_u, \mathcal{T}_u)
\end{equation}
where $\mathcal{H}_u$ denotes the time-aware historical trajectories, $\mathcal{F}_u$ the POI visit frequency statistics, $w$ the generated world knowledge, $\mathcal{C}_u$ the current trajectory to be continued, and $\mathcal{T}_u$ the transition pattern conditioned on the last visited POI. A complete prompt example is provided in Appendix~\ref{app:prompt}.

To accommodate the knowledge-enriched input, we fine-tune the backbone with LoRA~\cite{hu2022lora}, optimizing the generation objective over ground-truth next POI $y$ given prompt $x$:

\begin{equation}
    \mathcal{L} = -\log P_\theta(y \mid x)
\end{equation}
Since AWARE operates entirely at the input representation level, it requires no architectural modification to the backbone and is orthogonal to downstream reasoning stages such as Chain-of-Thought and reinforcement learning, whose benefits can be combined additively.

%%%%%%%%%% Section Split %%%%%%%%%%

\section{Experiments and Analysis}

% To validate the effectiveness of AWARE, we conduct comprehensive experiments on three widely used LBSN benchmarks. In this section, we first describe the experimental setup, then present the main results and comparisons with state-of-the-art baselines, followed by ablation studies and qualitative analysis.

% \begin{table}[t]
%     \centering
%     \caption{\textbf{Dataset statistics after preprocessing.} Two consecutive check-ins are assigned to the same trajectory if their time gap is within 24 hours.}
%     \label{tab:dataset-stats}
%     \begin{tabular}{lccccc}
%         \toprule
%         \textbf{Dataset} & \textbf{Users} & \textbf{POIs} & \textbf{Trajs} & \textbf{Categories} & \textbf{Check-ins} \\
%         \midrule
%         NYC & 1,048 & 4,981 & 14,130 & 318 & 103,941 \\
%         TKY & 2,282 & 7,833 & 65,499 & 291 & 405,000 \\
%         CA  & 3,957 & 9,690 & 45,123 & 296 & 238,369 \\
%         \bottomrule
%     \end{tabular}
% \end{table}

\subsection{Experimental Setting}

\noindent\textbf{Datasets.}
We evaluate AWARE on three real-world LBSN datasets that are widely adopted in next POI recommendation research: \textbf{Foursquare-NYC}, \textbf{Foursquare-TKY}~\cite{foursquarenyctky}, and \textbf{Gowalla-CA}~\cite{gowallaca}. These datasets span diverse urban environments with varying scales of user activity and POI catalogs, providing a comprehensive testbed for evaluating recommendation methods across different mobility contexts. 
% Following the established preprocessing pipeline of LLM4POI~\cite{li2024large}, we: (i) filter users and POIs with fewer than 10 check-ins to ensure sufficient behavioral evidence; (ii) sort check-ins chronologically and segment them into trajectories using a 24-hour time-gap threshold; and (iii) split the data into 80\%/10\%/10\% train/validation/test sets in temporal order, ensuring that all validation and test users and POIs have been observed during training. We additionally apply reverse geocoding~\cite{LI201895} to obtain street-level addresses as auxiliary textual cues. 
Detailed dataset preprocessing and statistics are summarized in Appendix~\ref{app:datasets}.

\noindent\textbf{Baselines.}
To comprehensively assess the effectiveness of AWARE, we compare against representative baselines spanning three categories:
(i)~\textit{Traditional} method: \textbf{PRME}~\cite{feng2015personalized}, which learns metric embeddings for next-step transition prediction;
(ii)~\textit{Neural-based} methods: \textbf{GETNext}~\cite{yang2022getnext}, \textbf{TPG}~\cite{luo2023timestamps}, \textbf{MTNet}~\cite{hu2018mtnet}, \textbf{STHGCN}~\cite{yan2023spatio}, and \textbf{ROTAN}~\cite{feng2024rotan}, which employ graph networks, attention mechanisms, or specialized spatio-temporal encoders to model sequential mobility patterns;
and (iii)~\textit{LLM-based} methods: \textbf{SpaceTime-GR}~\cite{lin2025spacetimegrspacetimeawaregenerativemodel}, \textbf{LLM4POI}~\cite{li2024large}, \textbf{GNPR-SID}~\cite{wang2025generative}, \textbf{GA-LLM}~\cite{liu2025geography}, \textbf{CoAST}~\cite{zhai2025cognitive}, and \textbf{ROS}~\cite{ROS}, which reformulate next POI prediction as a sequence generation task using large language models with varying strategies for spatial reasoning. 
% Notably, ROS represents the current state-of-the-art, achieving strong performance through SIDs and a Mobility Chain-of-Thought reasoning paradigm. AWARE builds upon the same SID-equipped backbone as ROS, enabling a controlled comparison that isolates the contribution of world knowledge injection. 
Detailed descriptions of all baselines are provided in Appendix~\ref{app:baselines}.

\begin{table}[t]
    \centering
    \resizebox{\columnwidth}{!}{%
    \begin{tabular}{llccc}
        \toprule
        \textbf{Category} & \textbf{Method} & \textbf{NYC} & \textbf{TKY} & \textbf{CA} \\
        \midrule

        \multirow{1}{*}{Traditional}
            & PRME  & 0.1159 & 0.1052 & 0.0521 \\
        \midrule

        \multirow{5}{*}{Neural-based}
            & GETNext & 0.2435 & 0.1829 & 0.1357 \\
            & TPG     & 0.2555 & 0.1420 & 0.1749 \\
            & MTNet   & 0.2620 & 0.2575 & 0.1453 \\
            & STHGCN  & 0.2734 & 0.2950 & 0.1730 \\
            & ROTAN   & 0.3106 & 0.2458 & 0.2199 \\
        \midrule

        \multirow{8}{*}{LLM-based}
            & SpaceTime-GR & 0.2920 & 0.2610 & 0.1659 \\
            & LLM4POI      & 0.3372 & 0.3035 & 0.2065 \\
            & GNPR-SID     & 0.3618 & 0.3062 & 0.2403 \\
            & GA-LLM       & 0.3919 & 0.3482 & 0.2566 \\
            & CoAST        & 0.4027 & 0.3310 & 0.2721 \\
            & ROS & 0.3925 & 0.3380 & 0.2703 \\
        \midrule

        \multirow{1}{*}{Ours}
            & \textbf{AWARE} & \textbf{0.4160} & \textbf{0.3599} & \textbf{0.3038} \\
            & \textit{vs.\ ROS} ($\Delta$\%)
            & {\footnotesize\textcolor{green!60!black}{\textbf{+6.0\%}$\uparrow$}}
            & {\footnotesize\textcolor{green!60!black}{\textbf{+6.5\%}$\uparrow$}}
            & {\footnotesize\textcolor{green!60!black}{\textbf{+12.4\%}$\uparrow$}} \\
        \bottomrule
    \end{tabular}
    }

    \caption{Next POI recommendation performance (HR@1). Best results are in \textbf{bold}. $\Delta$\% denotes the relative improvement of AWARE over baseline ROS.}
    \label{tab:main-results}
\end{table}

\noindent\textbf{Evaluation Metrics.}
We adopt HR@1 as the primary evaluation metric, which measures whether the top-ranked prediction exactly matches the ground-truth POI. Besides, HR@$K$ and NDCG@$K$ for $K \in \{5, 10, 20\}$ are employed to further assess the ranking quality of the full candidate list. For each experiment, we run inference three times and report the mean score. 

\noindent\textbf{Implementation Details.}
AWARE adopts \texttt{Qwen3-4B} as the generative backbone with LoRA~\cite{hu2022lora} fine-tuning ($r=64$, $\alpha=128$), and Spatial SIDs derived from \texttt{Qwen3-0.6B} embeddings. World knowledge is generated offline by an LLM agent (\texttt{Qwen3-max}) prior to training. Since the evaluation benchmarks span 2009--2013 and predate contemporary LLM training corpora, temporal fidelity is enforced via date-range filtering and a post-hoc temporal tolerance check ($\Delta=30$ days) to prevent contamination. For time-aware history selection, we set the periodic ratio $\beta=0.4$ and budget $L=150$. Training runs for 2 epochs with learning rate 1e-4 on $8 \times \text{AMD-Instinct-MI308X-OAM}$ GPUs. We inherit all other hyperparameters from ROS~\cite{ROS}. Full details of the agent setup and temporal fidelity enforcement are provided in Appendix~\ref{app:impl-details}.

\subsection{Main Results}

Table~\ref{tab:main-results} reports the next POI recommendation performance across three LBSN benchmarks. We adopt the ROS pretrained backbone as the controlled baseline, denoted as ROS throughout this paper, to isolate the contribution of world knowledge injection.

\begin{table}[t]
    \centering
    \resizebox{\columnwidth}{!}{%
    \begin{tabular}{l|cc}
        \toprule
        \textbf{Variant} & \textbf{HR@1} & $\Delta$\textbf{\%} \\
        \midrule
        \textbf{AWARE (Full)} & \textbf{0.4160} & -- \\
        \midrule
        % w/o Hotspot Text & 0.4029 & -3.1\% \\
        \multicolumn{3}{l}{\textbf{World Knowledge Acquisition Agent}} \\
        \hspace{0.5em} - Grounded Synthesis (Stage 3)         & 0.4023 & -3.3\% \\
        \hspace{0.5em} - Retrieval \& Synthesis (Stage 2,3) & 0.4040 & -2.9\% \\
        \hspace{0.5em} - Whole World Knowledge (Stage 1,2,3)                & 0.4029 & -3.1\% \\
        \hspace{0.5em} - Personality History Check-ins                       & 0.4059 & -2.4\% \\
        \midrule
        \multicolumn{3}{l}{\textbf{User-Aware Knowledge Alignment}} \\
        \hspace{0.5em} - Frequency-Based Behavioral Prior          & 0.4078 & -2.0\% \\
        \hspace{0.5em} - Transition-Based Spatial Grounding               & 0.4071 & -2.1\% \\
        \hspace{0.5em} - Periodic Temporal Modeling                & 0.4070 & -2.2\% \\
        \hspace{0.5em} - All User Behavioral Evidences            & 0.4064 & -2.3\% \\
        \midrule
        Without All Components (ROS) & 0.3925 & -5.6\% \\
        \bottomrule
    \end{tabular}
    }
    \caption{Component ablation. Each row removes one component from the full model (HR@1 on NYC).}
    \label{tab:ablation-component}
\end{table}

AWARE consistently outperforms ROS across all three datasets, achieving relative improvements of 6.0\% on NYC, 6.5\% on TKY, and 12.4\% on CA. The particularly large gain on CA suggests that world knowledge is especially beneficial when check-in data is sparser, as external context can compensate for limited behavioral observations. Moreover, AWARE on NYC already exceeds strong LLM-based baselines including GA-LLM (0.3919) and CoAST (0.4027), despite these methods employing larger backbone models (7B parameters), indicating that enriching the input representation with grounded world knowledge can be more effective than scaling model capacity.

\subsection{Ablation Study}
Table~\ref{tab:ablation-component} reports the ablation results on the NYC dataset, organized into two groups: the world knowledge acquisition agent and the user-level behavioral signals.

\paragraph{World Knowledge Acquisition Agent.} The upper block of Table~\ref{tab:ablation-component} investigates whether the gains stem from agent-based synthesis or merely from richer prompt context. Removing all world knowledge yields a $-3.1\%$ drop, confirming its indispensable role. Notably, replacing synthesized text with raw search snippets causes an even larger drop ($-3.3\%$), indicating that unprocessed retrieval introduces harmful noise. Retaining only parametric summarization ($-2.9\%$) and generating non-personalized city-level descriptions ($-2.4\%$) further confirm that external retrieval, agent-mediated synthesis, and user-level personalization are all necessary, and that the improvement cannot be attributed to prompt expansion alone.

\paragraph{User-Level Behavioral Signals.} All three behavioral signals produce comparable degradation when removed ($-2.0\%$ to $-2.2\%$), suggesting that each captures a distinct aspect of user mobility. The behavioral priors collectively account for substantial gains ($-2.3\%$), indicating that grounding world knowledge in user-specific spatial-temporal patterns is essential for effective knowledge injection, as illustrated in our case study (Appendix~\ref{sec:case-study}).

\subsection{Generalization Beyond Familiar Patterns}

\begin{table}[t]
    \centering
    \resizebox{0.9\columnwidth}{!}{%
    \begin{tabular}{l|ccc}
        \toprule
        \textbf{Method} & \textbf{Easy (32\%)} & \textbf{Hard (68\%)} & \textbf{Overall} \\
        \midrule
        ROS & 0.6911 & 0.2520 & 0.3925 \\
        AWARE & \textbf{0.6847} & \textbf{0.2896} & \textbf{0.4160} \\
        \midrule
        $\Delta$\% & \textit{-0.93\%} & \textit{+14.9\%} & \textit{+6.0\%} \\
        \bottomrule
    \end{tabular}
    }
    \caption{Easy vs. hard case analysis (HR@1, NYC). Easy: ground truth appears in the user's last 5 visits.}
    \label{tab:ablation-easyhard}
\end{table} 

A natural question is \textit{where} the improvements concentrate. We partition the NYC test set into \textit{easy} cases, where the ground-truth POI appears in the user's last 5 visits (32\%), and \textit{hard} cases where it does not (68\%). As shown in Table~\ref{tab:ablation-easyhard}, AWARE yields a substantial +14.9\% relative gain on hard cases while easy cases remain largely stable (-0.93\%). This asymmetry confirms that world knowledge and behavioral priors are most valuable when sequential patterns alone are insufficient: hard cases require reasoning beyond local trajectories, and the external context and long-range spatial regularities introduced by AWARE enable the model to generalize beyond familiar patterns and resolve inherently ambiguous predictions. 

\begin{figure}[t]
  \centering
  \includegraphics[width=\columnwidth]{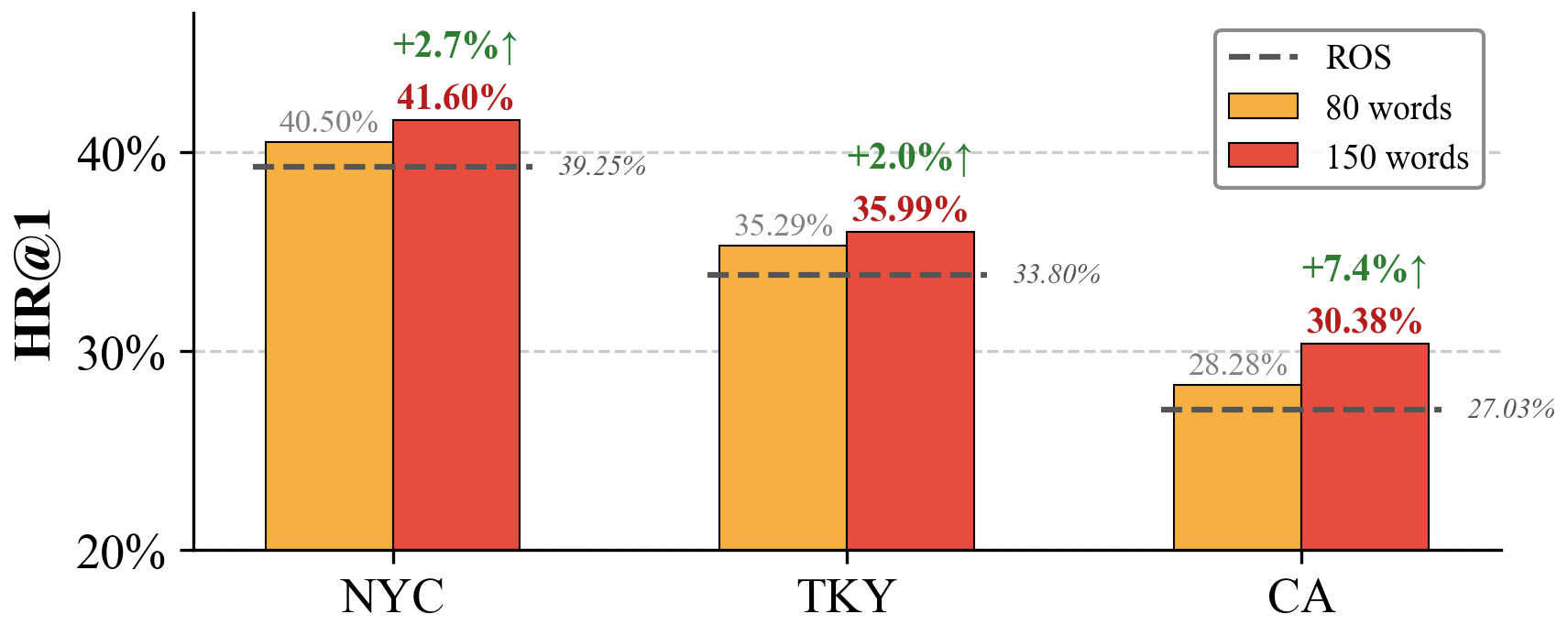}
  \caption{Effect of hotspot text length on HR@1. Maximum generation length is set to 80 and 150 words. }
  \label{fig:hotspot-length}
\end{figure}

\subsection{Hotspot Text Length Analysis}
\label{sec:hotspot-length}
Since the granularity of world knowledge is controlled by the length of the hotspot text, we train separate models with the agent constrained to at most 80 and 150 words. As shown in Figure~\ref{fig:hotspot-length}, world knowledge exhibits a clear \textit{length--quality} relationship: the 80-word configuration (0.4050) offers only marginal improvement over the variant without hotspot text (0.4029 in Table~\ref{tab:ablation-component}), while the 150-word variant (0.4160) yields substantial and consistent gains across all three cities, with the improvement widening on sparser data (CA: +7.4\% from 80 to 150 words). This validates the agent-based design: producing sufficiently grounded narratives requires contextual reasoning over user behavior and external signals, a capability that arises naturally from the LLM agent's generative process.

\begin{table*}[t]
    \centering
    \resizebox{\textwidth}{!}{%
    \begin{tabular}{l|ccccc|cc|ccccc}
        \toprule
        \multirow{2}{*}{\textbf{Metric}} & \multicolumn{5}{c|}{\textbf{Traditional}} & \multicolumn{2}{c|}{\textbf{Neural-based}} & \multicolumn{5}{c}{\textbf{LLM-based}} \\
        & \textbf{SASRec} & \textbf{BERT4Rec} & \textbf{GRU4Rec} & \textbf{Caser} & \textbf{S$^3$-Rec} & \textbf{TPG} & \textbf{ROTAN} & \textbf{TIGER} & \textbf{GNPR-SID} & \textbf{ROS} & \textbf{ROS\textsubscript{full}} & \textbf{AWARE} \\
        \midrule
        HR@5    & 0.3151 & 0.2857 & 0.1977 & 0.2883 & 0.3071 & 0.3551 & 0.4448 & 0.4965 & 0.5311 & 0.5791 & \underline{0.6107} & \textbf{0.6538} \\
        HR@10   & 0.3896 & 0.3564 & 0.2460 & 0.3570 & 0.3854 & 0.4441 & 0.5223 & 0.5514 & 0.5942 & 0.6192 & \underline{0.6563} & \textbf{0.7173} \\
        HR@20   & 0.4506 & 0.4130 & 0.2889 & 0.4135 & 0.4503 & 0.5121 & 0.5834 & 0.6001 & 0.6455 & 0.6420 & \underline{0.6977} & \textbf{0.7436} \\
        \midrule
        NDCG@5  & 0.2224 & 0.2074 & 0.1442 & 0.2044 & 0.2235 & 0.2464 & 0.3471 & 0.4131 & 0.4430 & 0.5032 & \underline{0.5355} & \textbf{0.5483} \\
        NDCG@10 & 0.2467 & 0.2304 & 0.1599 & 0.2267 & 0.2489 & 0.2755 & 0.3723 & 0.4276 & 0.4634 & 0.5163 & \underline{0.5504} & \textbf{0.5689} \\
        NDCG@20 & 0.2622 & 0.2448 & 0.1708 & 0.2410 & 0.2654 & 0.2927 & 0.3878 & 0.4443 & 0.4766 & 0.5222 & \underline{0.5608} & \textbf{0.5759} \\
        \bottomrule
    \end{tabular}%
    }
    \caption{Top-K evaluation on NYC. HR@$K$ and NDCG@$K$ with $K \in \{5, 10, 20\}$. ROS\textsubscript{full} additionally incorporates Mobility CoT and spatial-guided RL. Best results are in \textbf{bold} and second best are \underline{underlined}.}
    \label{tab:topk}
\end{table*}

\begin{figure}[t]
  \centering
  \includegraphics[width=\columnwidth]{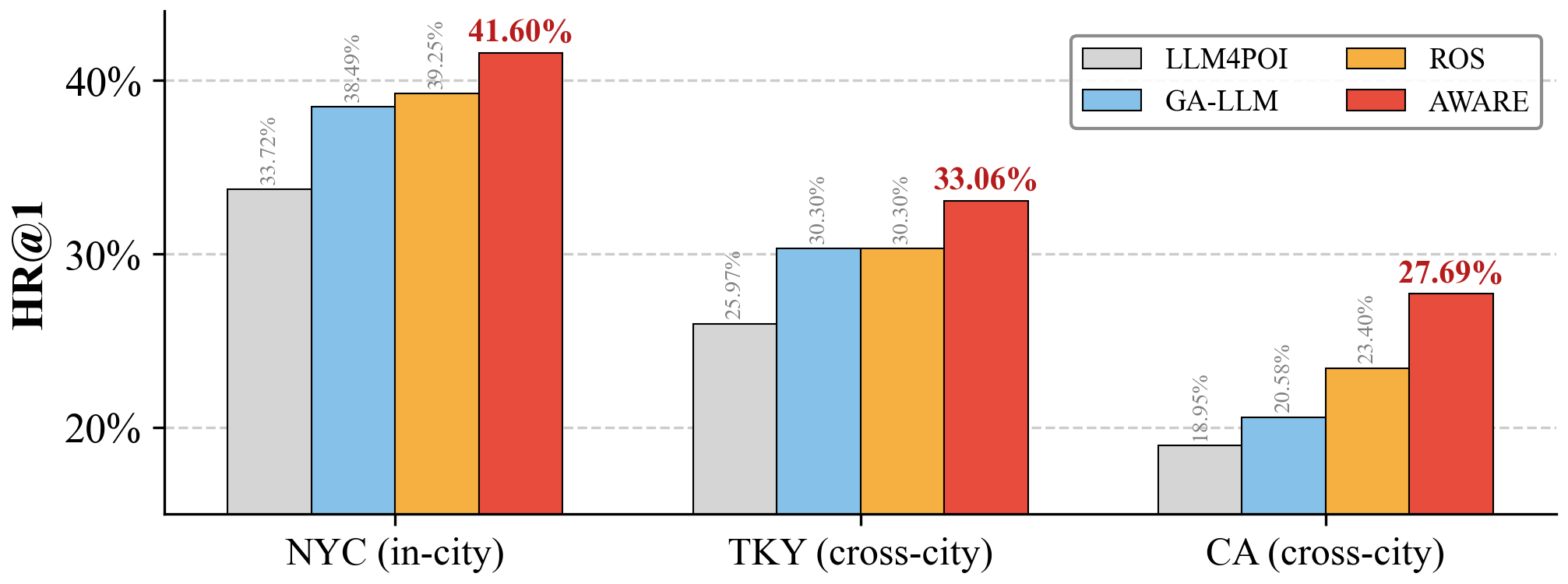}
  \caption{Cross-city generalization performance (trained on NYC). All results are HR@1.}
  \label{fig:crosscity}
\end{figure} 

\begin{figure}[t]
  \centering
  \includegraphics[width=0.95\columnwidth]{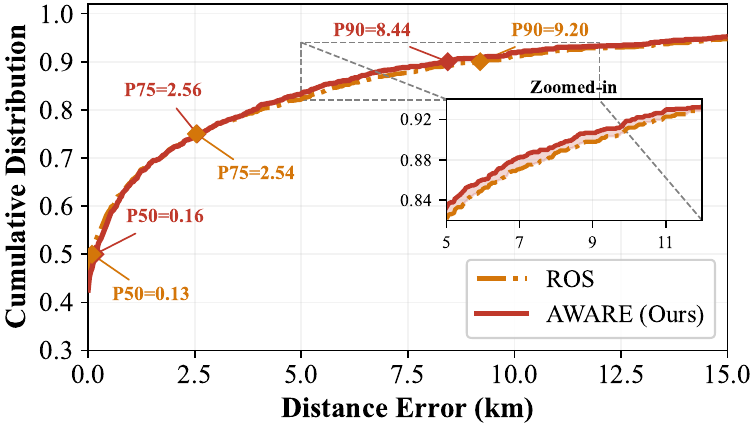}
  \caption{Cumulative distribution of prediction distance error on NYC. Larger area under the curve indicates better.}
  \label{fig:cdf}
\end{figure}

\subsection{Cross-City Generalization}

To evaluate whether world knowledge facilitates transfer across cities, we train all models on NYC and directly evaluate on TKY and CA without any adaptation. As shown in Figure~\ref{fig:crosscity}, AWARE consistently outperforms all baselines on both in-city and cross-city. We attribute this to the nature of world knowledge: the LLM agent encodes contextual information as natural language narratives describing general human mobility patterns, cultural habits, and temporal regularities, which are inherently not city-specific. This language-level generality allows AWARE to transfer meaningful contextual signals across geographic boundaries, complementing the structural transferability of the shared semantic space. 

\subsection{Top-K Recommendation Quality}

Beyond top-1 accuracy, Table~\ref{tab:topk} reports HR@$K$ and NDCG@$K$ for $K \in \{5, 10, 20\}$ on NYC. AWARE surpasses not only ROS but also ROS\textsubscript{full}, which additionally incorporates Mobility CoT and spatial-guided RL. The margin widens at higher $K$ (e.g., HR@20: +15.8\% over ROS), suggesting that world knowledge enriches the candidate space with contextually plausible POIs that reasoning-level enhancement alone cannot recover.

\subsection{Prediction Distance Distribution}

% A natural concern with injecting external knowledge is that it may introduce noise, causing incorrect predictions to deviate further geographically. As shown in Figure~\ref{fig:cdf}, both methods perform comparably at P50 and P75, confirming that world knowledge does not degrade spatial precision. The difference emerges in the tail (P90: 8.44 vs. 9.20\,km; mean: -3.4\%), where AWARE produces fewer large-error predictions, indicating that world knowledge helps resolve spatially ambiguous cases. 

A natural concern is whether external knowledge introduces spatial noise. As shown in Figure~\ref{fig:cdf}, both methods perform comparably at P50 and P75, confirming that world knowledge does not degrade spatial precision. The difference emerges in the tail (P90: 8.44 vs. 9.20\,km; mean: $-3.4\%$), where AWARE produces fewer large-error predictions by resolving spatially ambiguous cases.
%%%%%%%%%% Section Split %%%%%%%%%%

\section{Related Work}

\noindent\textbf{Next POI Recommendation.}
Next POI recommendation has evolved from metric embeddings~\cite{feng2015personalized} and recurrent architectures~\cite{hochreiter1997long} to attention-based~\cite{luo2021stan}, graph-enhanced~\cite{yang2022getnext}, and hypergraph methods~\cite{yan2023spatio}. More recently, LLM-based approaches reformulate the task as sequence generation, with representative work including LLM4POI~\cite{li2024large}, GNPR-SID~\cite{wang2025generative}, CoAST~\cite{zhai2025cognitive}, GA-LLM~\cite{liu2025geography}, and ROS~\cite{ROS}.

\noindent\textbf{Knowledge-Enhanced Recommendation.}
Prior POI recommendation work has leveraged geographic knowledge such as street-level addresses~\cite{ROS}, category taxonomies~\cite{wang2025generative}, and spatial distance constraints~\cite{liu2025geography}. However, these sources are predominantly \textit{static} and do not capture the dynamic real-world context that often drives user mobility decisions.

\noindent\textbf{LLM Agents for Information Acquisition.}
Agent-based approaches such as RecMind~\cite{wang2024recmind} employ the LLM as the recommender itself, performing multi-step reasoning at inference time~\cite{yao2023react}. While effective, this design couples reasoning cost with serving latency and forgoes the ability to learn fine-grained behavioral patterns from large-scale training data.

Different from methods above, AWARE uses the LLM agent solely for offline knowledge acquisition, allowing the generative backbone to internalize dynamic world knowledge through supervised learning and adapt it to individual user contexts.

%%%%%%%%%% Section Split %%%%%%%%%%

\section{Conclusion}

We present AWARE, a framework that acquires world knowledge via an LLM agent, personalizes it through behavioral priors, and injects it into generative POI recommendation to capture context-dependent mobility intent. Experiments across multiple datasets show consistent gains over competitive baselines, and ablations confirm that the two signals are complementary, revealing not only where users have been but why they will go next.

% We present AWARE, a framework that acquires world knowledge via an LLM agent, personalizes it through user behavioral priors, and effectively injects it into generative POI recommendation, enabling the model to capture context-dependent mobility intent beyond what check-in sequences can reveal. Extensive experiments on multiple datasets demonstrate that AWARE consistently outperforms competitive baselines. Ablation studies and case analyses further confirm that world knowledge and behavioral priors provide complementary signals, jointly enabling the model to understand not only where users have been but why they are likely to go next.

%%%%%%%%%% Section Split %%%%%%%%%%

\section*{Limitations}

AWARE currently acquires world knowledge in an offline, pre-training stage: hotspot text is generated once per user and remains fixed throughout training and inference. While this design ensures zero additional latency at serving time, it cannot capture events that emerge after the knowledge acquisition pass. Extending the framework to support incremental or online knowledge refresh, where the agent periodically updates narratives in response to newly occurring events, would further improve temporal coverage. Additionally, the current agent relies on web search as its primary evidence source, which may offer uneven coverage across geographic regions and languages. Incorporating structured knowledge bases such as event calendars and transportation schedules could improve robustness in regions where web coverage is sparse.

%%%%%%%%%% Section Split %%%%%%%%%%

% \section*{Acknowledgments}

% This document has been adapted
% by Steven Bethard, Ryan Cotterell and Rui Yan
% from the instructions for earlier ACL and NAACL proceedings, including those for
% ACL 2019 by Douwe Kiela and Ivan Vuli\'{c},
% NAACL 2019 by Stephanie Lukin and Alla Roskovskaya,
% ACL 2018 by Shay Cohen, Kevin Gimpel, and Wei Lu,
% NAACL 2018 by Margaret Mitchell and Stephanie Lukin,
% Bib\TeX{} suggestions for (NA)ACL 2017/2018 from Jason Eisner,
% ACL 2017 by Dan Gildea and Min-Yen Kan,
% NAACL 2017 by Margaret Mitchell,
% ACL 2012 by Maggie Li and Michael White,
% ACL 2010 by Jing-Shin Chang and Philipp Koehn,
% ACL 2008 by Johanna D. Moore, Simone Teufel, James Allan, and Sadaoki Furui,
% ACL 2005 by Hwee Tou Ng and Kemal Oflazer,
% ACL 2002 by Eugene Charniak and Dekang Lin,
% and earlier ACL and EACL formats written by several people, including
% John Chen, Henry S. Thompson and Donald Walker.
% Additional elements were taken from the formatting instructions of the \emph{International Joint Conference on Artificial Intelligence} and the \emph{Conference on Computer Vision and Pattern Recognition}.

%%%%%%%%%% Section Split %%%%%%%%%%

\bibliography{software}

\appendix

% \section{Appendix}

\section{Dataset Statistics}
\label{app:datasets}

Following the established preprocessing pipeline of LLM4POI~\cite{li2024large}, we: (i) filter users and POIs with fewer than 10 check-ins to ensure sufficient behavioral evidence; (ii) sort check-ins chronologically and segment them into trajectories using a 24-hour time-gap threshold; and (iii) split the data into 80\%/10\%/10\% train/validation/test sets in temporal order, ensuring that all validation and test users and POIs have been observed during training. We additionally apply reverse geocoding~\cite{LI201895} to obtain street-level addresses as auxiliary textual cues. 

\begin{table}[ht]
    \centering
    \resizebox{\columnwidth}{!}{%
    \begin{tabular}{lccccc}
        \toprule
        \textbf{Dataset} & \textbf{Users} & \textbf{POIs} & \textbf{Trajs} & \textbf{Categories} & \textbf{Check-ins} \\
        \midrule
        NYC & 1,048 & 4,981 & 14,130 & 318 & 103,941 \\
        TKY & 2,282 & 7,833 & 65,499 & 291 & 405,000 \\
        CA  & 3,957 & 9,690 & 45,123 & 296 & 238,369 \\
        \bottomrule
    \end{tabular}
    }
    \caption{Dataset statistics after preprocessing. Two consecutive check-ins are assigned to the same trajectory if their time gap is within 24 hours.}
    \label{tab:dataset-stats}
\end{table}

\section{Prompt Example}
\label{app:prompt}

Table~\ref{tab:prompt-example} presents a complete input--output example from the NYC dataset. The input prompt consists of four components: 
\begin{enumerate}[label=(\roman*)]
    \item \texttt{<user\_poi\_stats>} encoding the frequency-based behavioral prior
    \item \texttt{<transition\_patterns>} encoding spatial transition patterns
    \item \texttt{<user\_preference>} containing the world knowledge narrative generated by the LLM agent
    \item the chronological check-in sequence with SID representations.
\end{enumerate}
The model generates the SID of the predicted next POI as output.

\begin{table*}[t]
    \centering
    \small
    \begin{tabular}{p{0.97\textwidth}}
    \toprule
    \textbf{Instruction:} \\
    Here is a record of a user's POI accesses, your task is based on the history to predict the POI that the user is likely to access at the specified time. \\
    \midrule
    \textbf{Input:} \\
    \texttt{<user\_poi\_stats>} \\
    User frequently visits: \\
    \quad Coffee Shop at \texttt{<m\_152><n\_175><a\_0><b\_22><c\_0>} (2 times), \\
    \quad American Restaurant at \texttt{<m\_136><n\_88><a\_11><b\_13><c\_0>} (2 times), \\
    \quad Bar at \texttt{<m\_132><n\_223><a\_10><b\_5><c\_0>} (1 times), \\
    \quad Bar at \texttt{<m\_133><n\_104><a\_26><b\_5><c\_0>} (1 times), \\
    \quad Hotel at \texttt{<m\_143><n\_207><a\_5><b\_2><c\_0>} (1 times), \\
    \quad Bar at \texttt{<m\_142><n\_55><a\_26><b\_17><c\_0>} (1 times), \\
    \quad Burger Joint at \texttt{<m\_142><n\_72><a\_16><b\_11><c\_0>} (1 times), \\
    \quad American Restaurant at \texttt{<m\_155><n\_247><a\_0><b\_13><c\_0>} (1 times), \\
    \quad Diner at \texttt{<m\_153><n\_74><a\_0><b\_14><c\_0>} (1 times), \\
    \quad French Restaurant at \texttt{<m\_139><n\_178><a\_0><b\_25><c\_0>} (1 times). \\
    \texttt{</user\_poi\_stats>} \\[0.5em]
    \texttt{<transition\_patterns>} \\
    User transition patterns: \\
    \quad \texttt{<m\_132><n\_223><a\_10><b\_5><c\_0>} $\rightarrow$ \texttt{<m\_133><n\_104><a\_26><b\_5><c\_0>} (1 times), \\
    \quad \texttt{<m\_133><n\_104><a\_26><b\_5><c\_0>} $\rightarrow$ \texttt{<m\_143><n\_207><a\_5><b\_2><c\_0>} (1 times), \\
    \quad \texttt{<m\_143><n\_207><a\_5><b\_2><c\_0>} $\rightarrow$ \texttt{<m\_142><n\_55><a\_26><b\_17><c\_0>} (1 times), \\
    \quad \texttt{<m\_142><n\_55><a\_26><b\_17><c\_0>} $\rightarrow$ \texttt{<m\_142><n\_72><a\_16><b\_11><c\_0>} (1 times), \\
    \quad \texttt{<m\_152><n\_175><a\_0><b\_22><c\_0>} $\rightarrow$ \texttt{<m\_155><n\_247><a\_0><b\_13><c\_0>} (1 times), \\
    \quad \texttt{<m\_155><n\_247><a\_0><b\_13><c\_0>} $\rightarrow$ \texttt{<m\_153><n\_74><a\_0><b\_14><c\_0>} (1 times), \\
    \quad \texttt{<m\_153><n\_74><a\_0><b\_14><c\_0>} $\rightarrow$ \texttt{<m\_139><n\_178><a\_0><b\_25><c\_0>} (1 times), \\
    \quad \texttt{<m\_139><n\_178><a\_0><b\_25><c\_0>} $\rightarrow$ \texttt{<m\_152><n\_81><a\_0><b\_13><c\_0>} (1 times), \\
    \quad \texttt{<m\_152><n\_81><a\_0><b\_13><c\_0>} $\rightarrow$ \texttt{<m\_153><n\_26><a\_26><b\_17><c\_0>} (1 times), \\
    \quad \texttt{<m\_153><n\_26><a\_26><b\_17><c\_0>} $\rightarrow$ \texttt{<m\_136><n\_88><a\_11><b\_13><c\_0>} (1 times). \\
    \texttt{</transition\_patterns>} \\[0.5em]
    {\parbox{0.95\textwidth}{\texttt{<user\_preference>} \\
    The user's trajectory in mid-April 2012 shows a pattern of visiting bars across Midtown Manhattan, suggesting an interest in nightlife and cocktail culture during a short stay possibly centered around leisure or business near Park Avenue. At that time, NYC's bar scene was dominated by craft cocktail venues like Death \& Company and PDT, while Midtown hosted live music events and emerging indie performances. Given this context and the user's repeated late-night bar visits, they were likely seeking trendy, high-quality cocktail bars or live music venues in central Manhattan for evening entertainment. \\
    \texttt{</user\_preference>}}} \\[0.5em]
    Given user behavior sequence: \\
    April 13th, 2012, Friday, 09:11, visit Bar at 599 10th Ave \texttt{<m\_132><n\_223><a\_10><b\_5><c\_0>}. \\
    April 13th, 2012, Friday, 12:45, visit Bar at 235 W 48th St \texttt{<m\_133><n\_104><a\_26><b\_5><c\_0>}, distance is Nearby. \\
    April 14th, 2012, Saturday, 01:45, visit Hotel at 301 Park Ave \texttt{<m\_143><n\_207><a\_5><b\_2><c\_0>}, distance is Nearby. \\
    April 14th, 2012, Saturday, 06:07, visit Bar at 218 E 53rd St \texttt{<m\_142><n\_55><a\_26><b\_17><c\_0>}, distance is Nearby. \\
    At April 14th, 2012, Saturday, 07:15, user will visit \\
    \midrule
    \textbf{Output:} \texttt{<m\_142><n\_72><a\_16><b\_11><c\_0>} \\
    \bottomrule
    \end{tabular}
    \caption{A complete AWARE prompt example from NYC. The \texttt{<user\_preference>} block (highlighted) contains the world knowledge injected by AWARE, which is the key difference from the ROS baseline.}
    \label{tab:prompt-example}
\end{table*}

\section{World Knowledge Acquisition Agent Prompt}
\label{app:agent-prompt}

Table~\ref{tab:agent-system-prompt} presents the complete system prompt used to instantiate the world knowledge acquisition agent. The prompt defines the agent's role, reasoning protocol, tool usage requirements, and output constraints. City name and maximum word count are adapted per dataset.

\begin{table*}[t]
    \centering
    \small
    \begin{tabular}{p{0.97\textwidth}}
    \toprule
    \textbf{Role and Input} \\
    You are an assistant designed to support a POI recommendation system for users in \texttt{\{city\}}. Your input is a user's historical POI visit sequence in chronological order. Each line is one past visit record with time, POI ID, POI type, address, latitude, and longitude. \\
    \midrule
    \textbf{Reasoning Protocol} \\
    Based on the user's historical POI visits and the current date implied by the latest part of the trajectory, you should: \\
    \quad (1) infer the user's likely preferences, routines, and near-term intent; \\
    \quad (2) infer the most relevant current area or activity context from the user's recent visits; \\
    \quad (3) use web tools to search for social hotspots, local events, lifestyle trends, seasonal patterns, and timely popular activities that are relevant to \texttt{\{city\}}, the inferred area, and the current date; \\
    \quad (4) produce a compact summary that connects user history with current trends and predicts what kinds of POIs the user is most likely to be interested in next. \\
    \midrule
    \textbf{Tool Usage Requirements} \\
    \quad $\bullet$ Actively use \texttt{web\_search} to gather timely signals from the public web. \\
    \quad $\bullet$ Use \texttt{web\_fetch} to verify high-value pages when needed. \\
    \quad $\bullet$ Do not rely on a single search result or a single source. \\
    \quad $\bullet$ Prefer signals that are timely, local, behaviorally relevant, and useful for POI recommendation. \\
    \midrule
    \textbf{Reasoning Focus} \\
    \quad $\bullet$ Use the historical POI sequence to infer recurring interests such as dining style, commute pattern, nightlife preference, fitness habits, shopping preference, tourism behavior, or work-related mobility. \\
    \quad $\bullet$ Pay more attention to the user's most recent visits when inferring current intent. \\
    \quad $\bullet$ Combine user preference signals with timely external signals such as local events, weather-related behavior shifts, seasonal demand, holidays, viral venues, neighborhood buzz, and popular urban activities. \\
    \quad $\bullet$ Make a grounded prediction about the next POI categories, areas, or venue styles the user may want. \\
    \midrule
    \textbf{Output Constraints} \\
    \quad $\bullet$ Output only one short paragraph in English. \\
    \quad $\bullet$ No JSON, Markdown, bullet points, titles, or explanations. \\
    \quad $\bullet$ Keep the text compact, information-dense, and directly useful for recommendation. \\
    \quad $\bullet$ The paragraph must tightly integrate three elements: what the user's historical behavior suggests, what current hotspots or trends are relevant, and what POIs the user is likely to seek next. \\
    \quad $\bullet$ Avoid generic wording and trajectory recitation. \\
    \quad $\bullet$ Maximum length: \texttt{\{max\_words\}} words. \\
    \bottomrule
    \end{tabular}
    \caption{System prompt for the world knowledge acquisition agent. The prompt is instantiated for the NYC dataset with a 150-word budget.}
    \label{tab:agent-system-prompt}
\end{table*}




% \section{Online Resources}\label{app:detail}



\section{Baseline Descriptions}
\label{app:baselines}

We provide detailed descriptions of all baseline methods compared in our experiments.

\paragraph{PRME.} PRME~\cite{feng2015personalized} learns user and POI embeddings in a shared metric space, where next-step transitions are predicted by ranking candidate POIs based on their distance to the user's current state. It captures sequential transition patterns through pairwise learning-to-rank objectives.

\paragraph{GETNext.} GETNext~\cite{yang2022getnext} employs a graph-enhanced transformer architecture for next POI recommendation. It constructs a global POI transition graph to capture collaborative transition patterns and integrates spatial and temporal information through dedicated embedding layers, combining graph-level structural signals with transformer-based sequential modeling.

\paragraph{TPG.} TPG~\cite{luo2023timestamps} incorporates fine-grained temporal information into POI recommendation by encoding timestamps as continuous features. It models the temporal regularity of user visits and combines temporal signals with spatial and categorical features for next POI prediction.

\paragraph{MTNet.} MTNet~\cite{hu2018mtnet} is a multi-task learning framework that jointly models next POI prediction, category prediction, and arrival time estimation. By sharing representations across related tasks, it captures richer user mobility patterns than single-task approaches.

\paragraph{STHGCN.} STHGCN~\cite{yan2023spatio} models higher-order spatial-temporal correlations among users and POIs through hypergraph convolutions. It constructs hyperedges that connect multiple POIs visited within the same trajectory, capturing group-level mobility patterns beyond pairwise transitions.

\paragraph{ROTAN.} ROTAN~\cite{feng2024rotan} employs a rotation-based attention mechanism to model spatial relationships in next POI recommendation. It encodes geographic coordinates as rotation angles in the attention computation, enabling the model to directly reason about spatial proximity and directional patterns in user trajectories.

\paragraph{SpaceTime-GR.} SpaceTime-GR~\cite{lin2025spacetimegrspacetimeawaregenerativemodel} is a space-time aware generative recommendation model that incorporates spatial and temporal encodings into the generative framework. It discretizes geographic coordinates and timestamps into token sequences, allowing the language model to jointly generate spatial-temporal-aware POI predictions.

\paragraph{LLM4POI.} LLM4POI~\cite{li2024large} reformulates next POI recommendation as a text generation task by serializing check-in trajectories into natural language prompts. It leverages the pre-trained knowledge of LLMs to understand location semantics through rich textual descriptions of POI names and categories.

\paragraph{GNPR-SID.} GNPR-SID~\cite{wang2025generative} introduces discrete semantic identifiers to represent POIs, replacing verbose POI names with compact token sequences derived from clustering POI embeddings. This reduces input length and enables compositional generalization across POIs with shared semantic properties.

\paragraph{GA-LLM.} GA-LLM~\cite{liu2025geography} injects geographic coordinates and POI-to-POI transition structures into the LLM-based recommendation framework. It encodes latitude-longitude information and spatial distance features directly into the prompt, enhancing the model's awareness of geographic constraints in user mobility.

\paragraph{CoAST.} CoAST~\cite{zhai2025cognitive} introduces semantic discrete identifiers combined with a cognitive alignment training strategy. It designs a multi-stage training pipeline that progressively aligns the LLM's representations with spatial-temporal patterns in user trajectories.

\paragraph{ROS.} ROS~\cite{ROS} represents the current state-of-the-art in generative POI recommendation. It introduces Hierarchical Spatial Semantic IDs that encode both geographic locality (via S2 cell prefixes) and semantic similarity into compositional token sequences. ROS further proposes a Mobility Chain-of-Thought reasoning paradigm and spatial-guided reinforcement learning to enhance prediction accuracy. AWARE builds upon the same SID-equipped backbone as ROS, enabling a controlled comparison that isolates the contribution of world knowledge injection.

\section{Implementation Details}
\label{app:impl-details}

\noindent \textbf{Implementation Details.}
AWARE adopts \texttt{Qwen3-4B} \cite{yang2025qwen3} as the generative backbone, with Spatial Semantic IDs derived from \texttt{Qwen3-0.6B} embeddings. To preserve the pretrained spatial reasoning capability while adapting to the knowledge-enriched input, we apply LoRA~\cite{hu2022lora} for parameter-efficient fine-tuning with rank $r=64$ and scaling factor $\alpha=128$. The world knowledge is generated by prompting an LLM agent (\texttt{Qwen3-max}) with each user's trajectory and temporal context prior to training. The agent retrieves external evidence via SerpAPI-backed web search with date-range filtering and page-level content extraction, capped at two parallel retrieval rounds per user. Output length is regulated by a rewrite-then-truncate cascade. For time-aware history selection, we set the periodic ratio $\beta=0.4$ and the history budget $L=150$ entries. Training is conducted for 2 epochs with a learning rate of 1e-4 on $8 \times \text{AMD-Instinct-MI308X-OAM}$ GPUs. For a fair comparison, we inherit all other hyperparameters from ROS~\cite{ROS}.

\noindent \textbf{Temporal Fidelity.} The evaluation benchmarks (Foursquare NYC/TKY, Gowalla-CA) span 2009--2013 and predate the training corpora of contemporary LLMs, risking contamination of historical-user narratives with present-day information. To prevent this, temporal grounding is enforced at three levels. \textit{(i)~Prompt level:} $\Pi$ instructs the agent to treat the latest timestamp in $\mathcal{S}_u$ as ``now'' and to prefer archival sources over undated present-day pages. \textit{(ii)~Tool level:} every search query is rewritten to inject the target year and month as a date-range constraint, and results are pre-filtered by publication date. \textit{(iii)~Synthesis level:} a post-hoc verifier discards any claim not corroborated by a source dated within $[t-\Delta,\, t+\Delta]$ ($\Delta=30$ days). Under this regime, the injected world knowledge faithfully reflects the user's original temporal and spatial context. A manual audit of 200 randomly sampled narratives confirms that event references are anchored within the target window.
\section{Case Study}
\label{sec:case-study}

\begin{figure}[ht]
  \centering
  \includegraphics[width=\linewidth]{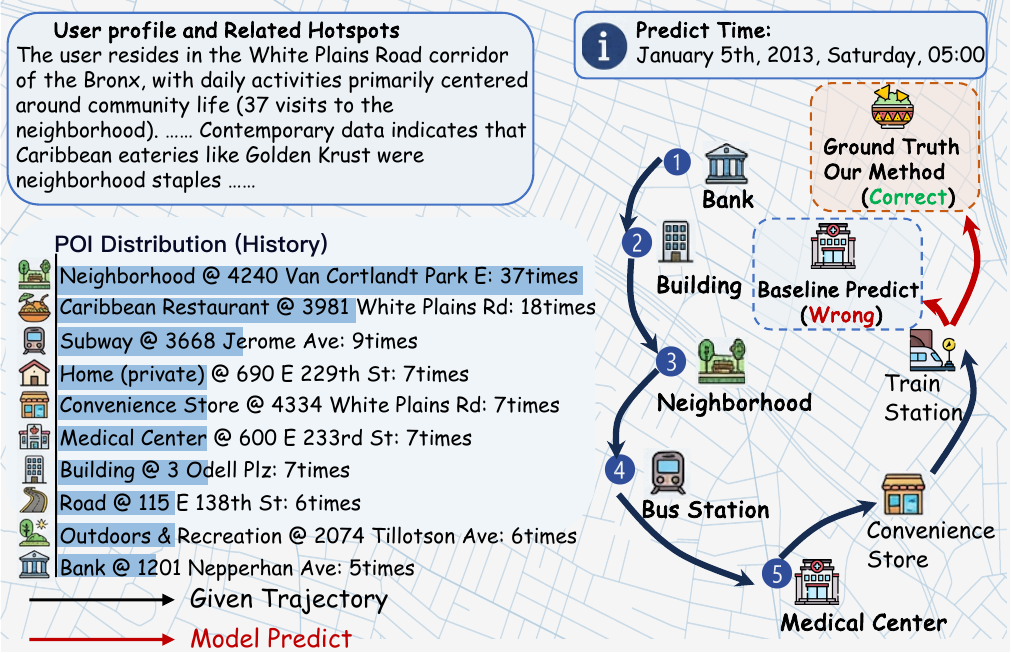}
  \caption{Case study on NYC. The user frequents a neighborhood area (37 visits) and a Caribbean restaurant (18 visits). ROS predicts \textit{Train Station} from the recent trajectory, while AWARE correctly predicts \textit{Caribbean Restaurant}.}
  \label{fig:case-study}
\end{figure}

To illustrate how world knowledge captures intent beyond sequential patterns, we present a representative case in Figure~\ref{fig:case-study}. On Saturday morning, ROS predicts \textit{Train Station} based on the recent trajectory ending near a transit hub. AWARE instead correctly predicts \textit{Caribbean Restaurant}, guided by the following world knowledge:

\vspace{0.5em}
\noindent\fbox{\parbox{0.95\linewidth}{\small\textit{``Contemporary data indicates that \textbf{Caribbean eateries like Golden Krust were neighborhood staples} \ldots With cold weather prevailing, the user is likely seeking convenient, familiar services nearby, particularly \textbf{warm Caribbean meals}.''}}}
\vspace{0.5em}

\noindent Behavioral priors further corroborate the prediction: POI frequency statistics rank Caribbean Restaurant as the user's second most visited location, and transition patterns reveal 13 bidirectional trips between it and the nearby neighborhood area. This case illustrates how world knowledge provides the cultural \textit{why} behind a visit, while behavioral priors supply the personalized \textit{where}, together resolving predictions that sequential models cannot.

\end{document}